\documentclass[letterpaper, 10 pt, conference]{IEEEtran}  % Comment this line out if you need a4paper

\IEEEoverridecommandlockouts                              % This command is only needed if 
\usepackage{graphics} % for pdf, bitmapped graphics files
\usepackage{epsfig} % for postscript graphics files
\usepackage{mathptmx} % assumes new font selection scheme installed
\usepackage{times} % assumes new font selection scheme installed
\usepackage{amsmath} % assumes amsmath package installed
\usepackage{amssymb}  % assumes amsmath package installed
\usepackage{cite}
\usepackage{float}
\usepackage{placeins}
\usepackage{textcomp}
\usepackage{graphicx}
\usepackage{subcaption}
\usepackage{color}
\usepackage{epstopdf}
\usepackage[linesnumbered,ruled,vlined]{algorithm2e}
\usepackage[left=0.75in,top=0.78in,right=0.75in,bottom=0.75in]{geometry}

\title{\LARGE \bf
Autonomous Robotic System using Non-Destructive Evaluation methods for Bridge Deck Inspection
}

\author{Tuan~Le,
	Spencer~Gibb,
	Nhan~Pham,
	Hung~Manh~La,~\IEEEmembership{Senior Member,~IEEE,}
	Logan~Falk,
	and~Tony~Berendsen%
\thanks{This work is supported by the University of Nevada, Reno and National Science Foundation under grants NSF-IIP\#1559942 and NSF-IIP\#1639092.}%
\thanks{Tuan Le, Spencer Gibb, Nhan Pham, Dr. Hung La and Logan Falk are with the Advanced Robotics and Automation (ARA) Lab, Department of Computer Science and Engineering, University of Nevada, Reno, NV, 89557, USA}.% (e-mails:  tuandzung.le, sgibb, nhan.ph@nevada.unr.edu, hla@unr.edu).}%
\thanks{ Tony Berendsen  is with Department of Mechanical Engineering, University of Nevada, Reno, NV, 89557, USA}% (e-mail: fberendsen@unr.edu, lfalk@nevada.unr.edu).}%
\thanks{The first two authors have made equal contributions to this work.}%
\thanks{\emph{Corresponding author:} Dr. Hung La (e-mail: hla@unr.edu).}%
}

\begin{document}

\maketitle
\thispagestyle{empty}
\pagestyle{empty}

%%%%%%%%%%%%%%%%%%%%%%%%%%%%%%%%%%%%%%%%%%%%%%%%%%%%%%%%%%%%%%%%%%%%%%%%%%%%%%%%

\begin{abstract}

Bridge condition assessment is important to maintain the quality of highway roads for public transport. Bridge deterioration with time is inevitable due to aging material, environmental wear and in some cases, inadequate maintenance. Non-destructive evaluation (NDE) methods are preferred for condition assessment for bridges, concrete buildings, and other civil structures. Some examples of NDE methods are ground penetrating radar (GPR), acoustic emission, and electrical resistivity (ER).  NDE methods provide the ability to inspect a structure without causing any damage to the structure in the process. In addition, NDE methods typically cost less than other methods, since they do not require inspection sites to be evacuated prior to inspection, which greatly reduces the cost of safety related issues during the inspection process. In this paper, an autonomous robotic system equipped with three different NDE sensors is presented. The system employs  GPR,  ER, and a camera for data collection. The system is capable of performing real-time, cost-effective bridge deck inspection, and is comprised of a mechanical robot design and machine learning and pattern recognition methods for automated steel rebar picking to provide realtime condition maps of the corrosive deck environments.              

\end{abstract}

%%%%%%%%%%%%%%%%%%%%%%%%%%%%%%%%%%%%%%%%%%%%%%%%%%%%%%%%%%%%%%%%%%%%%%%%%%%%%%%%
\section{Introduction}

Bridge deck condition assessment is the most important part for bridge health maintenance. The Federal Highway Administration (FHWA) initiated the Long-Term Bridge Performance (LTBP) program to utilize non-destructive evaluation (NDE) technologies for bridge deck condition assessment \cite{Gucunski15}. In a recent report, the number of concrete highway bridges in the United States with wearing surfaces is over 180,000 \cite{FHWA15}. Those bridges are prone to corrosion and without proper inspection, costly maintenances are inevitable. 
\begin{figure}
\centering
      	\includegraphics[width=\columnwidth]{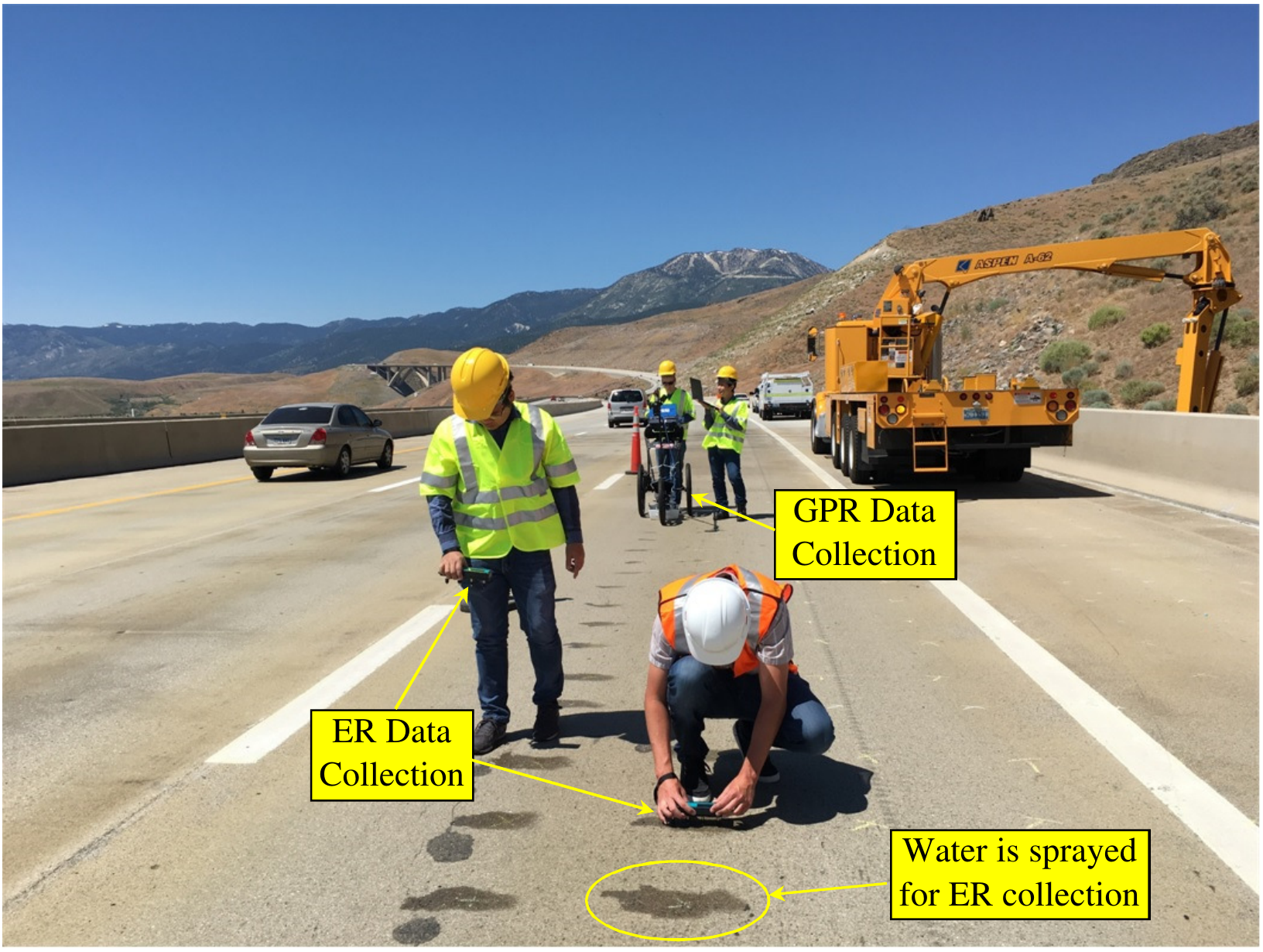}		
	\caption{Operation of NDE sensors by team members from the Advanced Robotics and Automation (ARA) Lab for bridge deck inspection on the Pleasant Valley Bridge on Highway 580 from Reno, NV toward Carson City, NV, July 2016.}
	\label{fig:data_collect_1}
	\vspace{-15pt}
\end{figure}
Even though a wide variety of NDE sensors are actively used in the field, the inspection process is still time-consuming, requiring skilled inspectors \cite{6723726,6743881,7292641}. As illustrated in Fig. \ref{fig:data_collect_1}, several inspectors were required to operate each sensor, which brings into question the safety requirements for inspection and the cost-efficiency of the inspection process. Furthermore, since sensor operation was conducted separately, discrepancies in collected data are possible. It will be beneficial if  a fully autonomous bridge deck inspection system is developed to address all of these issues, by eliminating the need for costly safety requirements, and reducing the number of paid hours spent on each inspection. There are several research efforts for the development of  the automated bridge inspection systems such as Robotic Bridge Inspection Tool (RABIT) \cite{Lim_ICRA2011,gucunski2013robotic,la2013mechatronic,la2014autonomous,6705706, Pham_CCC} associated with NDE data processing \cite{gucunski2015delamination,La_VE2015, dinha2015attenuation, gucunski2015implementation}. 

In this paper, a novel robotic system is designed for bridge deck inspection. The system is equipped with a digital single-lens reflex (DSLR) camera for visual crack detection, ground penetrating radar (GPR) for concrete rebar assessment, and two electrical resistivity (ER) sensors for concrete corrosion assessment. This system is designed to perform a comprehensive inspection of bridge decks. The camera system is used for bridge deck visual assessment and surface crack detection, while the GPR provides an in-depth look at the condition of the bridge and the steel rebar objects inside the concrete deck, which is the most vital part of a bridge. The ER sensors provide additional information about concrete deck condition in terms of the resistance of the concrete on the bridge deck, which tells how corroded the bridge deck is. The information collected from multiple sensors is integrated and being automatically processed to produce a comprehensive report of the bridge deck condition map. In addition, we propose a new method using machine learning for rebar detection. In comparison with previous work, this work is distinguished by providing a more cost-effective solution for autonomous bridge deck inspection, as well as showing that the proposed method for rebar picking works as well as the industry standard, and that it can perform in real-time by running it on multiple sets of real bridge data \cite{6653886}. 
\begin{figure*}[!t]
      	\includegraphics[width=\textwidth]{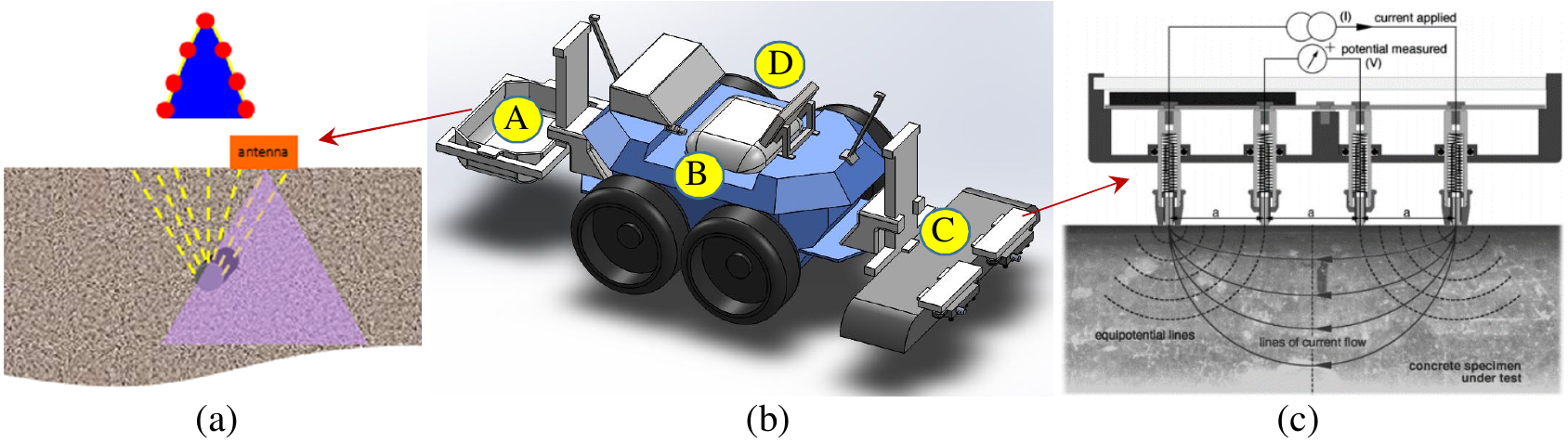}		
	\caption{A complete system design. (a) How GPR works; (b) An overall design A: GPR deployment system, B: Robot main body, C: ER deployment system, D: GPR display/monitor; (c) How ER works.}
	\label{fig:robot_design}
	\vspace{-0pt}
\end{figure*}
\begin{figure}
\centering
      	\includegraphics[width=\columnwidth, height = 4.9cm]{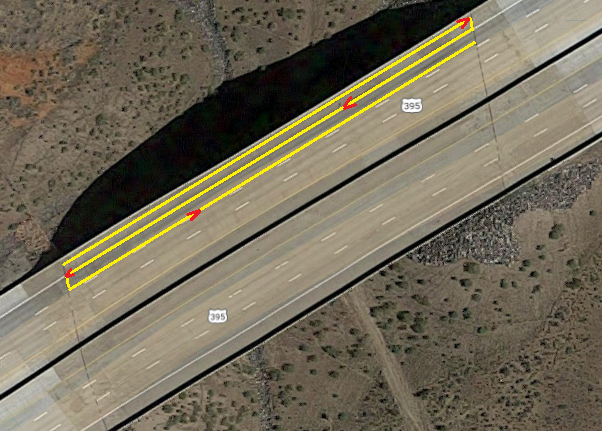}		
	\caption{Sample movement of a robot on a bridge deck. The robot moves along the yellow line in the direction of red arrows while collecting data.}
	\label{fig:Movement}
	\vspace{-10pt}
\end{figure}

The structure of the rest of this paper is as follows: in Section II, we discuss about the overall design of the robotic bridge deck inspection. In Section III, a new method for steel rebar detection is proposed. In Section IV, the system's capabilities are demonstrated through the experimental results.

\section{Overall Design For Robotic Bridge Deck Inspection}
In this section, the development of the robotic platform is described, and short explanations are provided to detail how NDE sensors work.
\subsection{Seekur Jr mobile robot as a base platform}
To have a robot move effectively along a narrow bridge deck, a skid-steering 4-wheel-drive robot model such as Seekur Jr mobile robot (from Omron Adept Technologies, Inc.) is used. To collect data along a bridge deck, the robot needs to move from one end of the bridge to the other end. Then it needs to turn around and continue its movement until the whole bridge deck is covered. The sample movement is illustrated in Fig. \ref{fig:Movement}. The selected mobile platform is able to rotate in place to change its movement direction, so that it can minimize unnecessary movements to conserve power. Besides, the Seekur Jr mobile robot is a waterproof platform, hence it is suitable for this bridge inspection task. The advantage of using this robot is that it is more versatile than its bigger version, Seekur, which had been used in other robotic systems \cite{6653886}. With a smaller form, the Seekur Jr. robot can manage to move in narrower environments. 

\subsection{Sensor integration}
\subsubsection{GPR}
GPR had been used in civil engineering for the last two decades \cite{maser1996condition,saarenketo2000road}. One of its applications is to evaluate corrosive level of top rebars inside bridge concrete decks \cite{dinha2015attenuation}. 

By sending a radar signal into the bridge deck and recording the two way travel time of the signal's reflection off of objects, it can produce unique signatures of that object; rebar in this case.
In Fig. \ref{fig:robot_design}(a), as the antenna moves past the buried rebar, it constructs the hyperbola signature of the rebar (red dot) continuously by recording the reflections (yellow lines) from the rebar. 
To attach the GPR to the robot, a deployment system was designed. 

In Fig. \ref{fig:robot_design}(b), a 3D model of a GPR deployment system is illustrated. The purpose of this deployment system is that the GPR unit will be touching the ground only when data is being collected. This design would make the robotic system easier to transport between locations.

\begin{figure*}[!t]
   	\includegraphics[width=\textwidth]{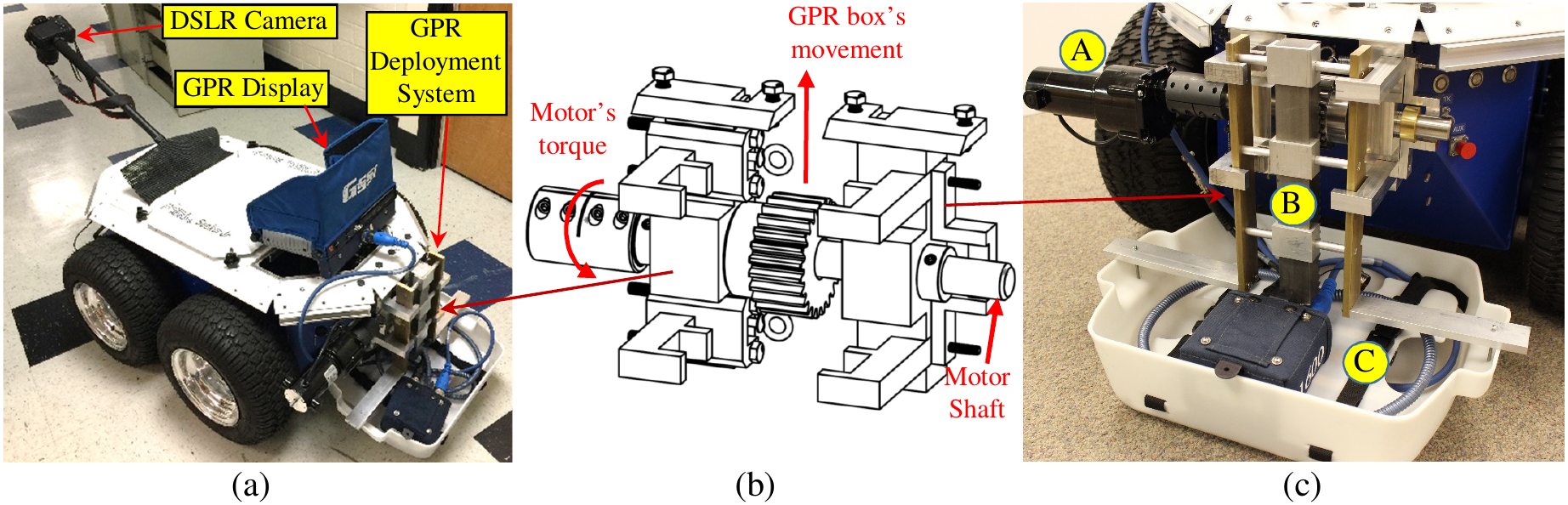}		
	\caption{Autonomous robotic system for bridge deck inspection: (a) System overview; (b) Schematic of GPR deployment system;  (c) Details of GPR deployment system: A - Motor, B - Gear shaft, C - GPR box.}
	\label{fig:whole_1}
	\vspace{-10pt}
\end{figure*}
%\begin{figure}[!htb]
%	\begin{subfigure}{8.8cm}
%		\includegraphics[height=5cm, width=8.8cm]{images/scan_selection_tab.png}
%		\caption{}
%		\label{fig:helena}
%	\end{subfigure}
%
%	\begin{subfigure}{8.8cm}
%		\includegraphics[height=5cm, width=8.8cm]{images/gpr_er_data_tab.png}
%		\caption{}
%		\label{fig:kendall}
%	\end{subfigure}

%	\begin{subfigure}{8.8cm}
%		\includegraphics[height=5cm, width=8.8cm]{images/camera_data _tab.png}
%		\caption{}
%		\label{fig:maine}
%	\end{subfigure}
%	\caption{Graphical user interface: (a) ``Scan Selection" tab; (b) ``GPR/ER Data" tab; (c) ``Camera Data" tab.}\label{fig:clahe}
%\vspace{-20pt}
%\end{figure}

\subsubsection{ER}
In Fig. \ref{fig:robot_design}(c), a diagram of how a Resipod ER works is presented. The selected ER sensor is from Proceq USA, Inc. The ER sensor measures the electrical resistance of the concrete of a bridge deck. Since there is a correlation between concrete deterioration and its resistance (i.e., higher resistance better concrete), the inclusion of ER sensor would provide further information about concrete bridge deck corrosion. 

\subsection{System implementation}
 A complete 3D design of the robotic system is presented in Fig.\ref{fig:robot_design}(b). The robotic system was tested in simulation before being implemented. 

Then a physical implementation of the robot is shown in Fig. \ref{fig:whole_1}. On this system, we also included a DSLR camera system to capture the bridge deck surface, which can be further processed for crack detection. To utilize the robot's mobility, the GPR and ER deployment systems are coupled with gear shafts so that they are only deployed when necessary. The GPR deployment system is installed on the rear of the robot while the ER deployment system is mounted on the front. The motors provide torques and via motor's shafts they control movements of GPR's box and ER sensors.The GPR deployment system with detailed parts is in Fig. \ref{fig:whole_1}(b),(c).  The robot localization and navigation is based on EKF-based sensor fusion from GPS, innertial measurement unit (IMU) and wheel odometry data \cite{la2013mechatronic}.

In order to easily visualize the data being collected by the robot,  a graphical user interface (GUI) was implemented. The GUI consists of three tabs: the ``Scan Selection" tab, the ``GPR/ER Data" tab, and the ``Camera Data" tab. The ``Scan Selection" tab shows the user a small preview of each finished scan and allows the user to dynamically update the data displayed on the other two tabs by selecting a scan. This tab will also update as more scans are performed, meaning it offers the user real-time access to the data being collected by the robotic system. The ``GPR/ER Data" tab shows the user a larger representation of the currently selected scan, as well as an image containing red squares where rebar were automatically detected. The ``Camera Data" tab lets the user navigate through a gallery of images taken of the ground surface by the camera. This tab automatically updates as the robotic system collects more data. Due to limit of space, the figure of this GUI is not presented here.

%Screenshots of the GUI implemented for the autonomous robotic system can be seen in Fig. \ref{fig:clahe}.    

\section{Steel Rebar Detection For Bridge Deck Condition Assessment}

\subsection{Related work}
Research in automated object detection using GPR is a recent development. Prior to this research, there were two primary methods for detecting objects in GPR scan images: manual detection using the human eye, and using commercial software \cite{Marecos2015}. These two methods have respective issues. Manually detecting hundreds of rebar in an image is time consuming and requires training to correctly identify rebar, and using commercial software requires purchasing an expensive software license. The goal of research in automated rebar detection is to offer a third option that is less expensive and requires less time from the user. 

Recent research in this field utilizes support vector machines, gradient descent, and various other computationally intense methods for detecting rebar \cite{Kaur2015, Shaw2005, Al-Nuaimy2000, la2014autonomous, la2013mechatronic}. While the accuracy of some recent methods has been reasonable, there is a recurring problem that methods are tested on data that is ideal or simulated \cite{Zhao2011, Pasolli2009}. However, the automated rebar detection method proposed in this paper combines image processing, image classification, and statistical methods to perform an otherwise time consuming task in real-time, while maintaining high accuracy and precision. 

\subsection{Algorithm}
The automated rebar detection method proposed in this paper utilizes contrast limited adaptive histogram equalization (CLAHE) to contrast stretch GPR scan images that are low contrast. This makes the rebar signatures more visible to the classifier and the human eye. Pixel intensity information, specifically the location of the black  horizontal area that indicates the presence of the ground plane,  is used to determine the location of the ground plane within GPR scan images, which functions as the start of the search area for the classifier. Edge detection is used to determine the average vertical location of the rebar in the image, which functions as the end of the search area for the classifier. Laplacian edge detection was used for this paper, but the type of edge detection has little bearing on the results of the method. 

A Naive Bayes classifier is used in this paper, since it has been shown to work well for simple classification tasks \cite{Shi2011}. The classifier classifies vectors of histogram of oriented gradients (HOG) features that are extracted from 50 by 15 pixel images. HOG features are used because they are invariant to geometric transformations and illumination, and they can quickly be computed. For these reasons, HOG features are preferable to other types of features and have been widely used in computer vision for object detection since they were first described by Dalal and Triggs in 2005 \cite{Dalal2005, Skibbe2012,  Nigam2013}. Information on the method for extracting a HOG feature vector from an image can be found in \cite{Dalal2005}.

%A HOG feature vector is extracted by first performing global normalization on a grayscale image: $I_N = \sqrt{ I}$. Then first order gradients of the normalized image are computed as $	\nabla{I_N} ={\begin{bmatrix}g_x \  g_y \end{bmatrix}}^T$, where $g_x$ is the normalized image gradient in the $x$ direction and $g_y$ is the normalized image gradient in the $y$ direction. The magnitude of the normalized image is given by $magnitude(I_N) = \sqrt{g_x^2 + g_y^2}$ and the orientation of the normalized image is given by $\theta = tan_2^-1(g_y, (g_x + 1*10^{-15})) * (180/\pi) + 90$. The image is then segmented into cells. It was empirically determined that 5 by 5 pixel cells yielded the best results for the proposed method. For each of the cells in the image, a histogram of gradient orientations is accumulated using the previously computed image information. Then, block normalization is applied. Blocks consist of several cells, and 3 by 3 cell blocks were used for this paper. Block normalization increases the invariance of HOG features by normalizing across a larger region of the image. 

\begin{figure}
	\centering
	\begin{subfigure}{0.92cm}
		\includegraphics{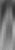}
		\caption{}
		\label{fig:h1}
	\end{subfigure}
	\hspace{0.1cm}
	\begin{subfigure}{0.92cm}
		\includegraphics{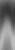}
		\caption{}
		\label{fig:h2}
	\end{subfigure}
	\hspace{0.1cm}
	\begin{subfigure}{0.92cm}
		\includegraphics{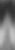}
		\caption{}
		\label{fig:h3}
	\end{subfigure}
	\hspace{0.1cm}
	\begin{subfigure}{0.92cm}
		\includegraphics{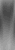}
		\caption{}
		\label{fig:nh1}
	\end{subfigure}
	\hspace{0.1cm}
	\begin{subfigure}{0.92cm}
		\includegraphics{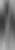}
		\caption{}
		\label{fig:nh2}
	\end{subfigure}
	\hspace{0.1cm}
	\begin{subfigure}{0.92cm}
		\includegraphics{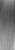}
		\caption{}
		\label{fig:nh3}
	\end{subfigure}
	%\par\bigskip
	\caption{(a)-(c) Positive samples used in the training process that contain clear hyperbolas indicating the presence of rebar; (d)-(f) Negative samples used in the training process that do not contain a hyperbola. Each of these 6 samples is 50 by 15 pixels.}\label{fig:training}
\vspace{-15pt}
\end{figure}

The classifier is trained using the HOG feature vectors extracted from manually selected images that are assigned class labels, indicating that they either contain a clearly centered hyperbola (class 1) or don't (class 2). Information on the training data set can be seen in Table \ref{tab:training} and Fig. \ref{fig:training}. Once training is complete, a sliding window is applied across the GPR scan image, within the search window previously determined. At each sliding window location, a HOG feature vector is extracted from the window and the classifier determines if the window contains a hyperbola. Given a vector of HOG features, $x = (x_1, ...,x_n)$, where $n$ is the number of features in the vector, and Bayes' theorem, it holds that $p(C_k|x) = p(C_k)p(x|C_k)$, where $C_k$ is class $k$. This can be rewritten as $p(C_k|x) = p(C_k, x_1,...,x_n)$, where $k$ is the number of classes. Now, using the general product rule, this can be rewritten again as $p(C_k|x) = p(x_1|x_2,...,x_n,C_k)* p(x_2|x_3,...,x_n,C_k)...p(x_n-1,x_n,C_k) * p(x_n|C_k)p(C_k)$. By assuming that every feature in the feature vector is conditionally independent of the other features, the Bayes model can be written as $p(C_k|x)=\frac{1}{Z} p(C_k) \prod_{i=1}^{n}p(x_i|C_k)$. Using this model, class labels can be assigned to test samples using the equation: $\hat{y} = \operatorname*{arg\,max}_{k\in{\{1,...,K\}}}  p(C_k) \prod_{i=1}^{n}p(x_i|C_k)$, where $\hat{y}$ is the assigned class label given a sample, which is chosen based on the maximum probability of a class given the sample. 

\begin{table}
\caption{{\sc Training Data Set}}
\centering
\begin{tabular}{c c c}
\hline\hline 
Class & Class Name & Number of Images \\ [0.5ex] 
\hline  
1 & Hyperbolas & 304  \\ 
2 & Not Hyperbolas & 1800  \\ [1ex] 
\hline
\end{tabular}
\label{tab:training} 
\vspace{-10pt}
\end{table}

The classification process yields a set of points, $P_{OUT}$ indicating the general location of rebar within the GPR scan image. These points effectively function as clusters in image space where rebar are most likely to be. Areas with a large number of points are more likely to contain rebar than areas with few points. Because of the steps that are performed on the points from the classifier, it is not necessary for the classifier to be extremely accurate. This is advantageous because the Naive Bayes classifier can be trained quickly compared to its more complex alternatives like fuzzy classifiers and support vector machines.

Once the classifier has output a set of points indicating general rebar locations within the image, more precise localization must be done. Typical methods for hyperbola localization and fitting include the Hough transform or RANSAC algorithm \cite{Kaur2015, Al-Nuaimy2000}. However, these methods are time consuming, which makes them less than ideal for use in real-time systems. The method used in this paper is referred to as histogram localization, and uses statistics and pixel information to accurately localize rebar signatures in GPR scan images in real-time. 

Histogram localization takes three values as input: the set of points from the classifier, now referred to as $P_{IN}$, the start of the search location, $s$, and the end of the search location, $e$. The first step of this process entails a histogram of $x$ coordinates of each point being accumulated. Then, non-maxima suppresion is performed, leaving only the local maxima remaining in the histogram. Next, the highest intensity pixel along each maxima and within the search area are located. This yields a set of points. Finally, another search is performed in 5 by 5 pixel neighborhood around each point to determine where the highest intensity pixel value is. This is based on the assumption that rebar signatures are largely white, indicating the reflection of the radar off of a metalic object. A final non-maxima suppresion step can be performed to  ensure that only the top points in the image are kept, in the cases where multiple points may exist on a single rebar. A detailed description of these steps can be seen in Algorithm 1.

%%%%%%%%%%%%%%%%%%%%%%%%%%
% Algorithm 2
%%%%%%%%%%%%%%%%%%%%%%%%%%
\begin{algorithm}
\DontPrintSemicolon
\KwIn{$P_{IN} = \{P_1, P_2,...,P_n\} | P_n = (x_n, y_n) $\; 
$s$ = starting search location\;
$e$ = ending search location}
\KwOut{$P_{OUT} = \{P_1, P_2,..., P_3\}$}

$x\_histogram[Image\ width]$

\For{$x \gets P_{IN}[0][0]$ \KwTo $P_{IN}[n][0]$}{
$x\_histogram[x]$ += $1$\;
}
\For{$i \gets 0$ \KwTo $x\_histogram.length$}{
\If{$x\_histogram[i] > 0$}{
$maxima \gets true$\;

$Maxima\_list \gets [\ ]$\;

\For{$j \gets i - 7$ \KwTo $i+6$}{
\If{$j > -1\ and\ j < x\_histogram.length$}{
\If{$x\_histogram[j] > x\_histogram[i]$}{
$maxima \gets false$\;
}
}
}

\If{$maxima == true$}{
append $maxima$ to $Maxima\_list$\;
}
}
}
$x\_coords \gets [\  ]$\;
$y\_coords \gets [\  ]$\;

\For{$i \gets 0$ \KwTo $Maxima\_list.length$}{
$x \gets -1$\;
$y \gets -1$\;

\For{$j \gets search\_start$ \KwTo $search\_end$}{
\If{$Image[j, Maxima\_list[i]]>x$}{
$x \gets Image[j, Maxima\_list[i]]$\;
$y \gets j$\;
}
}

append $x$ to $x\_coords$\;
append $y$ to $y\_coords$\;
}
$P_{OUT} \gets [\ ]$\;

\For{$i \gets 0$ \KwTo $x\_coords.length$}{
$x \gets x\_coords[i]$\;
$y \gets y\_coords[i]$\;

$intensity \gets Image[x,y]$\;

$final\_x \gets -1$\;
$final\_y \gets -1$\;
\For{$j \gets y-3$ \KwTo $y+2$}{
\For{$k \gets x-3$ \KwTo $x+2$}{
\If{$Image[j,k] > intensity$}{
$intensity \gets Image[j,k]$\;
$final\_x \gets k$\;
$final\_y \gets j$\;
}
}
}
append $(final\_x, final\_y)$ to $P_{OUT}$ 
}

\caption{{\sc Precise Hyperbola Localization}}
\label{algorithm:LHL}
\vspace{-0pt}
\end{algorithm}
%%%%%%%%%%%%%%%%%%%%%%%%%%
The steps detailed in the algorithm have been tested on real bridge data that was collected for this research. The results show that the proposed method is able to accurately locate rebar within a GPR scan image, in real-time. Run time of the proposed method has been included in the results to show the ability of this system to operate efficiently in real-world situations where time is limited and accuracy is of the utmost importance. More information on results of this algorithm can be seen in Section IV.

\section{Experimental Results}

\begin{figure}[!h]
\centering
      	\includegraphics[width=\columnwidth, height = 5cm]{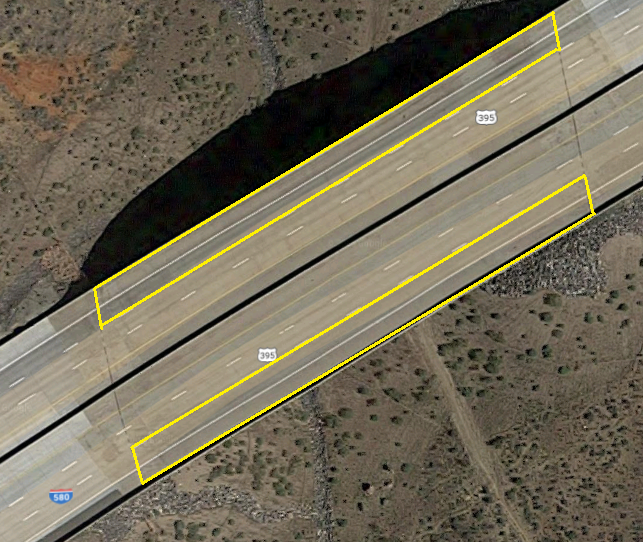}		
	\caption{Pleasant Valley Bridge on Highway 580, Reno, Nevada with the surveyed areas marked by yellow lines (image taken from Google Map).}
	\label{fig:Bridge_scan}
	\vspace{-10pt}
\end{figure}

To assess the efficiency of the proposed method, data collected from the Pleasant Valley Bridge, on Highway 580, Reno, Nevada is used. The slow lane and shoulder were surveyed on both sides of this bridge on July 28-29, 2016 during the day as marked by yellow lines in Fig. \ref{fig:Bridge_scan}. 

The rebar picking results from proposed method are compared with the rebar picking results from RADAN\textregistered 7. RADAN\textregistered 7 is produced by Geophysical Survey System\textregistered, Inc. (GSSI). The results concentrate on rebar locating ability and are presented in terms of accuracy, precision, number of true positives, and number of false positives. A true positive in this case is a rebar being detected. A false positive is a detection being made when there is no rebar present. Accuracy is the percent of rebar that are correctly identified; $true\_positives/total\_reba$r. Precision is the percent of identified objects that are true positives; $true\_positives/(true\_positives+false\_positives)$.
\setlength{\textfloatsep}{1\baselineskip plus 0.2\baselineskip minus 0.5\baselineskip}

In order to show that the proposed method performs well on multiple sets of real bridge data, it was tested on the  Pleasant Valley Bridge and East Helena Bridge, as well as on two additional bridges. Run time has been included in the results for these bridges to show that the proposed method is capable of processing this data in real-time. The results of the proposed method on the additional bridges can be seen in Table 1.  

The performance of the propsed method was compared to the performance of RADAN\textregistered 7. These results can be seen in Table 2. It can be seen that the proposed method performs better than RADAN\textregistered 7 on the data from the East Helena Bridge, and that the difference in performance between the two is negligable on the data from the Pleasant Valley Bridge. 
%For a visual representation of the performance comparison between the proposed method and RADAN\textregistered 7, see Fig. \ref{fig:results_comparison}.

After obtaining the rebar locations asscociated with their amplitude of reflected signal, the attenuation condition map of the deck is built. For example, the condition map of  the Pleasant Valley Bridge is presented in Fig. \ref{fig:condition_map} which provides a visual representation of the corrosive condition of the bridge deck. There are four corrosive levels: no corrosive - blue; low corrosive - green; moderate corrosive - orange; highly corrosive - red. As can be seen, this bridge is in good condition since there are only some very small red areas. Due to limit of space, results of deck image stitching, crack mapping and ER condition map are not presented here.

\begin{table*}
\centering
	\caption{Automated rebar detection results of the proposed method.}
	\begin{tabular}{l c c c c c r}
	\hline \hline
	Bridge name & Location & Images & Total rebar & Accuracy & Precision & Run time/image \\ 
	\hline
	Kendall Pond Road Bridge & Derry, NH & 12 & 2284 & 91.46\% & 97.79\% & 32.91s \\
	Ramp D Bridge & Lewiston, ME & 14 & 3699 & 92.89\% & 93.787\% & 55.46s \\
	\end{tabular}
\label{tab:results1}
\end{table*}

\begin{table*}
\centering
	\caption{Automated rebar detection comparison: proposed method vs. RADAN 7.}
	\begin{tabular}{l c c c c c c c r}
	\hline \hline
	Bridge name & Location & Images & Total rebar & Method &  True positives & False positives & Accuracy & Precision \\
	\hline
	Pleasant Valley Bridge & Reno, NV & 20 & 13205 & Proposed Method & 12768 & 52 & 96.69\% & 99.59\% \\
	 & & & & RADAN\textregistered 7 & 13135 & 186 & 99.47\% & 98.60\% \\
	East Helena Bridge & Helena, MT & 14 & 1055 & Proposed Method & 1046 & 19 & 99.15\% & 98.22\% \\
	 & & & & RADAN\textregistered 7 & 917 & 151 & 86.92\% & 85.86\% \\
	\end{tabular}
\label{tab:results2}
\vspace{-20pt}
\end{table*}
 
%\begin{figure}
%	\begin{subfigure}{\columnwidth}
%		\includegraphics[width=\columnwidth, height = 1.8cm]{images/4c.png}
%	\caption{}
%	\end{subfigure}
%	\par \bigskip
%	\begin{subfigure}{\columnwidth}
%		\includegraphics[width=\columnwidth, height = 1.8cm]{images/4radanc.png}
%	\caption{}
%	\end{subfigure}
%	\par \bigskip
%	\begin{subfigure}{\columnwidth}
%		\includegraphics[width=\columnwidth, height = 1.8cm]{images/16c.png}
%	\caption{}
%	\end{subfigure}
%	\par \bigskip
%	\begin{subfigure}{\columnwidth}
%		\includegraphics[width=\columnwidth, height = 1.8cm]{images/16radanc.png}
%	\caption{}
%	\end{subfigure}
%
%	\caption{(a) An example of the output from the proposed method, where some rebar are not detected (yellow squares); (b) An example of the output from RADAN\textregistered 7, where the rebar not detected in (a) are successfully detected; (c) An example of the output from the proposed method run on the East Helena Bridge, where no rebar are missed; (d) An example of the output from RADAN\textregistered 7 run on the East Helena Bridge, where rebar that were detected in (c) are missed (yellow squares). }
%\label{fig:results_comparison}
%\vspace{-10pt}
%\end{figure}

\begin{figure}
\centering
 	\includegraphics[width=\columnwidth]{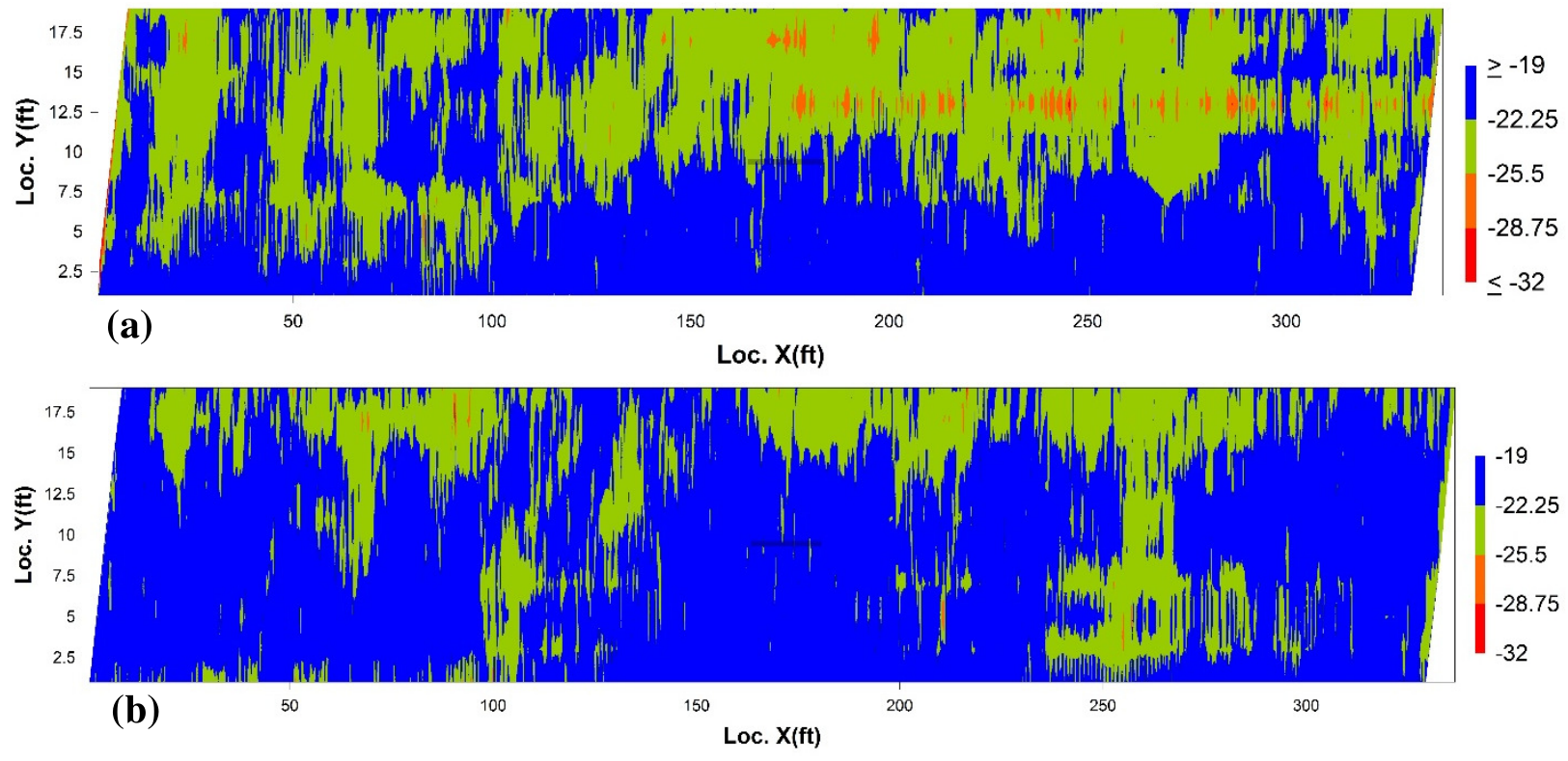}		
	\caption{Condition map of Pleasant Valley Bridge: (a) for the bottom marked area in Fig. \ref{fig:Bridge_scan}; (b) for the top marked area in Fig. \ref{fig:Bridge_scan} .}\label{fig:condition_map}
\label{fig:condition_map}
\vspace{-10pt}
\end{figure}

%To provide a visual assessment of the bridge deck, collected images are stitched together to creat an entire image for the ease of inspection as shown in Fig. \ref{fig:stitching}. As can be seen, the overall condition of the  Pleasant Valley Bridge deck is in good condition (no crack) that is consistant with the GPR condition map as shown in  Fig.   \ref{fig:condition_map}. Due to space limit, the bridge condition map from ER sensor is not presented in this paper.
%\begin{figure*}
%\centering
%      	\includegraphics[width=.8\textwidth]{Image_Stiching_HWY580.eps}		
%	\caption{Sample of image stitching for visual assessment of the Pleasant Valley Bridge deck: Top: zoom-in at some locations of the deck; Middle: stitched image; Bottom: individual images collected at every 2 feet (60cm).}
%	\label{fig:stitching}
%	\vspace{-10pt}
%\end{figure*}

\section{Conclusions}

In this paper, a novel and autonomous robotic system for bridge deck inspection has been developed. As new bridges are constantly being constructed and the condition of current bridges needs to be monitored effectively, this system can be widely applied to reduce bridge maintenance costs. Also, this system can be easily adapted by equipping it with various NDE sensors. Additionally, the proposed automated rebar detection method allows the robot to process the GPR data in realtime and generate the bridge deck condition map accurately. Future work on this system will include improved GUI integration of NDE sensors, additional invariance of the classifier to edge cases that may typically cause poor performance, and more testing on real bridge data. %Such functionality and testing will make it possible for the robotic system to operate in tough environments, such as multi-deck bridges, and allow it to perform bridge inspection almost anywhere. 

\section*{Acknowledgement}

The authors would like to acknowledge the Nevada Department of Transportation for their support and allowing us to access the bridge for system testing and data collection.

\bibliography{ref}
\bibliographystyle{IEEEtran}

\end{document}